# Visual Character Recognition using Artificial Neural Networks

*Shashank Araokar*[*]
*MGM's College of Engineering and Technology,*
*University of Mumbai, India*
*(shashank.araokar@ieee.org)*

### ABSTRACT

The recognition of optical characters is known to be one of the earliest applications of Artificial Neural Networks, which partially emulate human thinking in the domain of artificial intelligence. In this paper, a simplified neural approach to recognition of optical or visual characters is portrayed and discussed. The document is expected to serve as a resource for learners and amateur investigators in pattern recognition, neural networking and related disciplines.

## [1.] INTRODUCTION:

The recognition of characters from scanned images of documents has been a problem that has received much attention in the fields of image processing, pattern recognition and artificial intelligence. Classical methods in pattern recognition do not as such suffice for the recognition of visual characters due to the following reasons:

---

[*] Presently pursuing Bachelor's degree in Electronics and Telecommunication Engineering under the University of Mumbai, India

1. The 'same' characters differ in sizes, shapes and styles from person to person and even from time to time with the same person.
2. Like any image, visual characters are subject to spoilage due to noise.
3. There are no hard-and-fast rules that define the appearance of a visual character. Hence rules need to be heuristically deduced from samples.

As such, the human system of vision is excellent in the sense of the following qualities:

1. The human brain is adaptive to minor changes and errors in visual patterns. Thus we are able to read the handwritings of many people despite different styles of writing.
2. The human vision system learns from experience: Hence we are able to grasp newer styles and scripts with amazingly high speed.
3. The human vision system is immune to most variations of size, aspect ratio, color, location and orientation of visual characters.

In contrast to limitations of classical computing, Artificial Neural Networks (ANNs), that were first developed in the mid 1900's serve for the emulation of human thinking in computation to a meager, yet appreciable extent. Of the several fields wherein they have been applied, *humanoid computing* in general and *pattern recognition* in particular have been of increasing activity. The recognition of visual (optical) characters is a problem of relatively amenable complexity when compared with greater challenges such as recognition of human faces. ANNs have enjoyed considerable success in this area due to their humanoid qualities such as adapting to changes and learning from prior experience. The subsequent parts of the paper elucidate this fact in more details.

The paper is organized as follows: in section [2.], *image digitization*, which is an essential step prior to neural networking, is described. Section [3.] describes the learning mechanism of the neural network used, and the employed architecture is described in section [4.]. Section [5.] discusses the issues that affect the performance of the proposed methods with reference to its accuracy, computational complexity and extensibility.

[2.] IMAGE DIGITIZATION:

When a document is put to visual recognition, it is expected to be consisting of printed (or handwritten) characters pertaining to one or more *scripts* or *fonts*. This document however, may contain information besides optical characters alone. For example, it may contain pictures and colors that do not provide any useful information in the instant sense of character recognition. In addition, characters which need to be *singly* analyzed may exist as *word clusters* or may be located at various points in the document. Such an image is usually processed for noise-reduction and separation of individual characters from the document. It is convenient for comprehension to assume that the submitted image is freed from noise and that individual characters have already been located (using for example, a suitable clustering algorithm). This situation is synonymous to the one in which a single noise-free character has been submitted to the system for recognition.

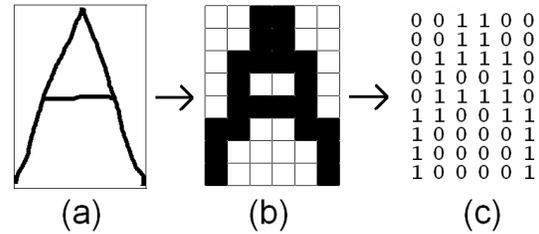

Fig. (1)

The process of digitization is important for the neural network used in the system. In this process, the input image is sampled into a binary window which forms the input to the recognition system. In the above figure, the alphabet *A* has been digitized into *6X8=48* digital cells, each having a single color, either black or white. It becomes important for us to encode this information in a form meaningful to a computer. For this, we assign a value *+1* to each black pixel and *0* to each white pixel and create the binary image matrix *I* which is shown in the Fig. (1.c). So much of conversion is enough for neural networking which is described next. Digitization of an image into a binary matrix of specified dimensions makes the input image invariant of its actual dimensions. Hence an image of whatever size gets transformed into a binary matrix of fixed pre-determined dimensions. This establishes uniformity in the dimensions of the input and stored patterns as they move through the recognition system.

[3.] LEARNING MECHANISM:

In the employed system, a highly simplified architecture of artificial neural networks is used. For purpose of easy understanding, the learning mechanism of the neural network is described first and its architecture is described next, in section [4.]. In the used method, various characters are *taught* to the network in a supervised manner. A character is presented to the system and is assigned a particular *label*. Several variant patterns of the same character are taught to the network under the same label. Hence the network learns various possible variations of a single pattern and becomes adaptive in nature. During the *training process*, the input to the neural network is the input matrix *M* defined as follows:

$If\ I(i, j) = 1\ Then\ M(i, j) = 1$
*Else:*
$If\ I(i, j) = 0\ Then\ M(i, j) = -1$  (1.1)

The input matrix *M* is now fed as input to the neural network. It is typical for any neural network to learn in a supervised or unsupervised manner by adjusting its *weights*. In the current method of learning, each candidate character taught to the network possesses a corresponding weight matrix. For the $k^{th}$ character to be taught to the network, the weight matrix is denoted by $W_k$. As learning of the character progresses, it is this weight matrix that is updated. At the commencement of teaching (supervised training), this matrix is initialized to zero. Whenever a character is to be taught to the network, an input pattern representing that character is submitted to the network. The network is then *instructed* to identify this pattern as, say, the $k^{th}$ character in a *knowledge base* of characters. That means that the pattern is assigned a label *k*. In accordance with this, the weight matrix $W_k$ is updated in the following manner:

```
for all i=1 to x
{
  for all j=1 to y
  {
    W_k(i, j) = W_k(i, j) + M(i, j)
  }
}
```
(1.2)

Here *x* and *y* are the dimensions of the matrix $W_k$ (and *M*).

The following figure shows the digitization of three input patterns representing *S* that are presented to the system for it to learn.

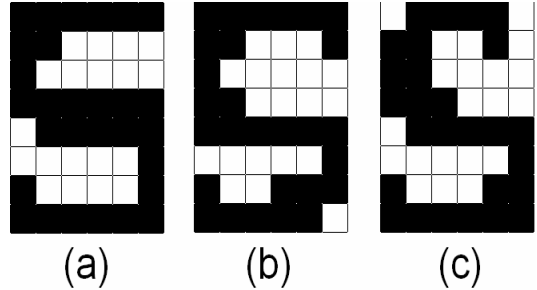

(a)     (b)     (c)
Fig. (2)

Note that the patterns slightly differ from each other, just as handwriting differs from person to person (or time to time) and like printed characters differ from machine to machine.

Fig. (3) gives the weight matrix, say, $W_S$ corresponding to the alphabet *S*. The matrix is has been updated thrice to learn the alphabet *S*. It should be noted that this matrix is specific to the alphabet *S* alone. Other characters shall each have a corresponding weight matrix.

$$W_s = \begin{pmatrix} 1 & 3 & 3 & 3 & 3 & 1 \\ 3 & 3 & -3 & -3 & -1 & -1 \\ 3 & -1 & -3 & -3 & -3 & -3 \\ 3 & 3 & 1 & -1 & -1 & -1 \\ -1 & 3 & 3 & 3 & 3 & 3 \\ -3 & -3 & -3 & -3 & -3 & 3 \\ 3 & -3 & -3 & -1 & 1 & 3 \\ 3 & 3 & 3 & 3 & 3 & 1 \end{pmatrix}$$

Fig. (3)

A close observation of the matrix would bring the following points to notice:

1. The matrix-elements with higher (positive) values are the ones which stand for the most commonly occurring image-pixels.
2. The elements with lesser or negative values stand for pixels which appear less frequently in the images.

Neural networks learn through such updating of their weights. Each time, the weights are adjusted in such a manner as to give an output closer to the desired output than before. The weights may represent the *importance* or *priority* of a parameter, which in the instant case is the occurrence of a particular pixel in a character pattern. It can be seen that the weights of the most frequent pixels are higher and usually positive and those of the uncommon ones are lower and often negative. The matrix therefore assigns importance to pixels on the basis of their frequency of occurrence in the pattern. In other words, highly probable pixels are assigned higher priority while the less-frequent ones are penalized. However, all labeled patterns are treated without bias, so as to include impartial *adaptation* in the system.

## [4.] NETWORK ARCHITECTURE:

The overall architecture of the recognition system is shown in Fig. (4). In this system, the candidate pattern *I* is the input. The block '*M*' provides the input matrix *M* to the weight blocks $W_k$ for each *k*. There are totally *n* weight-blocks for the totally *n* characters to be taught (or already taught) to the system.

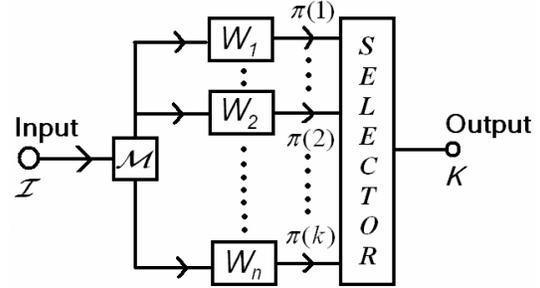

Fig. (4)

The recognition of patterns is now done on the basis of certain statistics that shall be defined next.

*(4.1) Candidate Score ($\psi$):* This statistic is a product of corresponding elements of the weight matrix $W_k$ of the $k^{th}$ learnt pattern and an input pattern *I* as its *candidate*. It is formulated as follows:

$$\psi(k) = \sum_{i=1}^{x} \sum_{j=1}^{y} W_k(i,j) * I(i,j) \qquad (1.3)$$

It should be noted that unlike in the training process where *M* was the processed input matrix, in the recognition process, the binary image matrix *I* is directly fed to the system for recognition.

*(4.2) Ideal Weight-Model Score ($\mu$):* This statistic simply gives the sum total of all the positive elements of the weight matrix of a learnt pattern. It may be formulated as follows (with $\mu(k)$ initialized to *0* each time).

*for i=1 to x*
*{*
  *for j=1 to y*
  *{*
    *if* $W_k(i, j) > 0$ *then*
    *{*
      $\mu(k) = \mu(k) + W_k(i, j)$
    *}*
  *}*
*}*         (1.4)

*(4.3) Recognition Quotient (Q):* This statistic gives a measure of how well the recognition system identifies an input pattern as a *matching* candidate for one of its many learnt patterns. It is simply given by:

$$Q(k) = \frac{\psi(k)}{\mu(k)} \quad (1.5)$$

The greater the value of *Q*, the more confidence does the system bestow on the input pattern as being similar to a pattern already known to it. The classification of input patterns now follows the following trivial procedure:-

1. For an input candidate pattern *I*, calculate the recognition quotient ($Q(k)$) for each learnt pattern *k*.
2. Determine the value of *k* for which $Q(k)$ has the maximum value.
3. Too low maximum value of $Q(k)$ (say less than 0.5) indicates poor recognition. In such a case:

- Conclude that the candidate pattern does not exist within the knowledge base OR
- Teach the candidate pattern to the network till a satisfactory value of $Q(k)$ is obtained.

4. Conditionally, identify the input candidate pattern as being akin to the $k^{th}$ learnt pattern OR proceed with the training for better performance.

In Fig. (4), the selector gives an output *k* by making the best selection as in Step 4 of the aforementioned algorithm. The adaptive performance of the network can easily be tested by an example: we submit two hand-drawn patterns representing *S* and *P* respectively to the system that has already learnt only the character *S*. The recognition quotient yielded by the trained system is mentioned alongside.

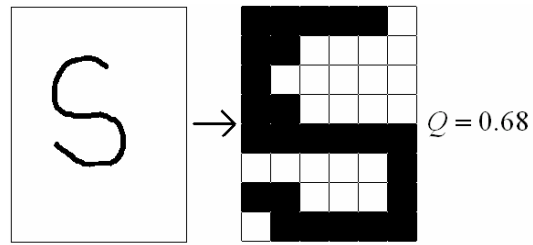

Fig. (5)

Note that the pattern in Fig. (5) does not exactly appear like the three patterns of Fig. (2) that were taught to the system. However, being adaptive, the system nevertheless bestows a good quotient $Q = 0.68$ on the pattern, indicating a match. To improve recognition of this particular pattern, the same pattern can be repeatedly input to the system and taught to it as before under the same label. As a result, the value of *Q* approaches unity after each time the pattern is taught. This illustrates *learning* *from prior experience* in neural networks.

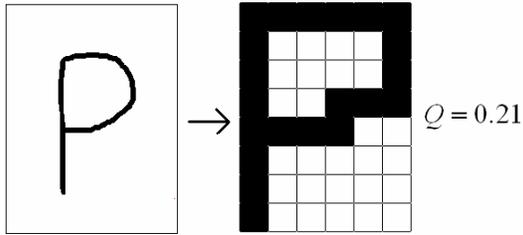

Fig. (6)

The system however dismisses the candidature of the pattern representing *P* in Fig. (5) by yielding a low value of $Q(=0.21)$. It can be observed by regular teaching, that the system develops on its ability to identify a matching pattern and reject non-matching patterns. Thus, regular supervised teaching marks enhanced performance of the system.

### [5.] PERFORMANCE ISSUES:

The neural system has some direct advantages that become apparent at this stage:

1. The method is highly adaptive; recognition is tolerant to minor errors and changes in patterns.
2. The knowledge base of the system can be modified by teaching it newer characters or teaching different variants of earlier characters.
3. The system is highly general and is invariant to size and aspect ratio.
4. The system can be made user-specific: User-profiles of characters can be maintained, and the system can be made to recognize them as per the orientation of the user.

The dimensions of the input matrix need to be adjusted for performance. Greater the dimensions, higher the resolution and better the recognition. This however increases the time-complexity of the system which can be a sensitive issue with slower computers. Typically, *32X32* matrices have been empirically found sufficient for the recognition of English handwritten characters. For intricate scripts, greater resolution of the matrices is required.

As already illustrated in the previous example, efficient supervised teaching is essential for the proper performance. Neural expert systems are therefore typically used where human-centered training is preferred against rigid and inflexible system-rules.

### [6.] CONCLUSION:

A simplistic approach for recognition of visual characters using artificial neural networks has been described. The advantages of neural computing over classical methods have been outlined. Despite the computational complexity involved, artificial neural networks offer several advantages in pattern recognition and classification in the sense of emulating adaptive human intelligence to a small extent.

### [7.] ACKNOWLEDGEMENT:

The author wishes to acknowledge the use of the Open Source VB-based software *Recog* developed by Neil Fraser for the purpose of character recognition using neural networks. Parts of this paper have been derived from the same program. It is available for download at the following web address: http://neil.fraser.name/software/recog

## [8.] REFERENCES: